\documentclass[10pt,twocolumn,letterpaper]{article}
\usepackage[pagenumbers]{cvpr} 
\usepackage{graphicx}
\usepackage{amsmath}
\usepackage{amssymb}
\usepackage{booktabs}
\usepackage{epigraph}
\usepackage{url}            
\usepackage{amsfonts}      
\usepackage{nicefrac}      
\usepackage{xcolor,colortbl}         
\usepackage{xspace}
\usepackage{multirow}
\usepackage{mathtools}
\usepackage{comment}
\usepackage{color}
\usepackage{subfiles}
\usepackage{bbding}
\usepackage{bbm}
\usepackage{dsfont}
\usepackage[pagebackref,breaklinks,colorlinks]{hyperref}
\usepackage[capitalize]{cleveref}
\DeclareMathOperator*{\argmax}{argmax}

\DeclareMathOperator*{\softmax}{softmax}
\crefname{section}{Sec.}{Secs.}
\Crefname{section}{Section}{Sections}
\Crefname{table}{Table}{Tables}
\crefname{table}{Tab.}{Tabs.}
\newcommand{\PAR}[1]{\vskip4pt \noindent {\bf #1~}}
\newcommand{\PARit}[1]{\vskip4pt \noindent {\it #1~}}

\newcommand*{\markgg}{\cellcolor{gray!15}}
\newcommand*{\markg}{\cellcolor{gray!2}}

\newcommand*{\marklgg}{\rowcolor{gray!15}}
\newcommand*{\marklg}{\rowcolor{gray!2}}

\begin{document}

\title{Pix2Map: Cross-modal Retrieval for Inferring Street Maps from Images}

\author{
    Xindi Wu $^{1}\thanks{Now at Princeton, work done while at Carnegie Mellon University.}$
    \qquad 
    KwunFung Lau$^{1}$\thanks{Now at Intel.}
    \qquad 
    Francesco Ferroni$^{2}$\thanks{Now at Nvidia.}
    \qquad 
    Aljoša Ošep$^{1}$
    \qquad 
    Deva Ramanan$^{1, 2}$  \\
    $^{1}$Carnegie Mellon University
    \quad 
    $^{2}$Argo AI \\
{\tt\small xindiw@princeton.edu, kwun.fung.lau@intel.com, fferroni@nvidia.com, \{aosep, deva\}@andrew.cmu.edu}\\
\small{\texttt{\url{pix2map.github.io}}}}

\maketitle

%%%%%%%%% ABSTRACT
\begin{abstract}
Self-driving vehicles rely on urban street maps for autonomous navigation. In this paper, we introduce \textit{Pix2Map}, a method for inferring urban street map topology directly from ego-view images, as needed to continually update and expand existing maps. This is a challenging task, as we need to infer a complex urban road topology directly from raw image data. The main insight of this paper is that this problem can be posed as cross-modal retrieval by learning a joint, cross-modal embedding space for images and existing maps, represented as discrete graphs that encode the topological layout of the visual surroundings. We conduct our experimental evaluation using the Argoverse dataset and show that it is indeed possible to accurately retrieve street maps corresponding to both seen and unseen roads solely from image data. Moreover, we show that our retrieved maps can be used to update or expand existing maps and even show proof-of-concept results for visual localization and image retrieval from spatial graphs. 

\end{abstract}

\section{Introduction}
\label{sec:intro}

We propose \textit{Pix2Map}, a method for inferring road maps directly from images. More precisely, given the camera images, \textit{Pix2Map} generates a topological map of the visible surroundings, represented as a spatial graph. Such maps encode both geometric and semantic scene information such as lane-level boundaries and locations of signs~\cite{Thrun2003RoboticMA} and serve as powerful \textit{priors} in virtually all autonomous vehicle stacks. In conjunction with on-the-fly sensory measurements from lidar or camera, such maps can be used for localization~\cite{barsan2020learning} and path planning~\cite{ma2019exploiting}. As map maintenance and expansion to novel areas are challenging and expensive, often requiring manual effort~\cite{liu2020high,seif2016autonomous}, automated map maintenance and expansion has been gaining interest in the community~\cite{lambert2021trust,liu2020high,liang2019convolutional,homayounfar2019dagmapper,li2021hdmapnet,mi2021hdmapgen,can2021structured,can2022topology}.

\PAR{Why is it hard?} To estimate urban street maps, we need to learn to map continuous images from ring cameras to discrete graphs with varying numbers of nodes and topology in bird's eye view (BEV). Prior works that estimate road topology from monocular images first process images using Convolutional Neural Networks or Transformers to extract road lanes and markings~\cite{can2021structured} or road centerlines~\cite{can2022topology} from images. These are used in conjunction with recurrent neural networks for the generation of polygonal structures~\cite{castrejon2017annotating} or heuristic post-processing~\cite{li2021hdmapnet}  to estimate a spatial graph in BEV. This is a very difficult learning problem: such methods need to jointly learn to estimate a non-linear mapping from image pixels to BEV, as well as to estimate the road layout and learn to generate a discrete spatial graph.

\begin{figure}[t!]
\centering
   \includegraphics[width=\linewidth]{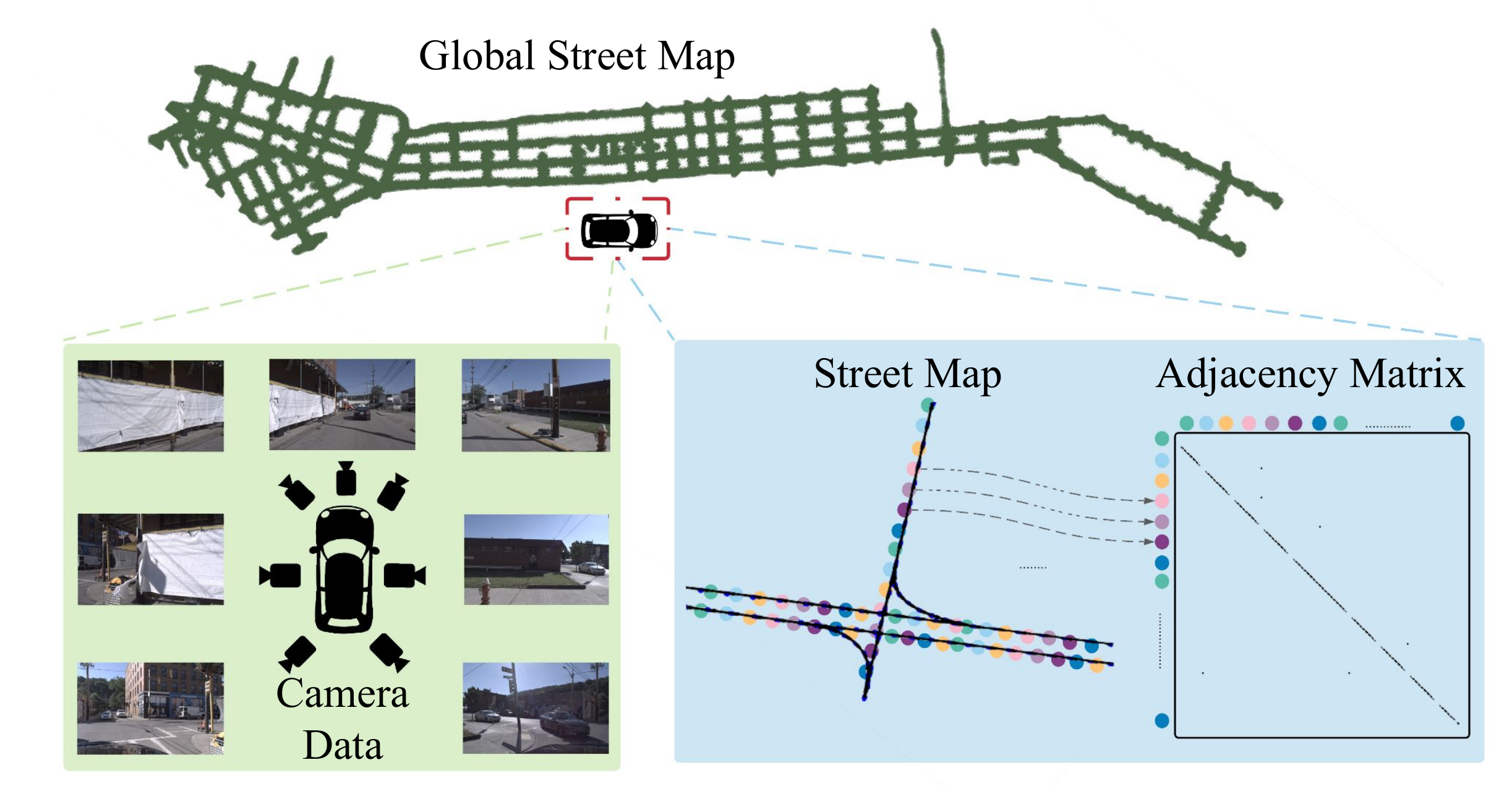}
  \caption{\textbf{Illustration of our proposed Pix2Map for cross-modal retrieval.} Given unseen 360$^{\circ}$ ego-view images collected from seven ring cameras (\textit{left}), \textit{Pix2Map} predicts the local street map by retrieving from the existing street map library (\textit{right}), represented as an adjacency matrix. The local street maps can be further used for global high-definition map maintenance (\textit{top}).
   }
\label{fig:encoder}
\end{figure}

\PAR{Pix2Map.} Instead, our core insight is to simply sidestep the problem of graph generation and 3D localization from monocular images by recasting \textit{Pix2Map} as a \textit{cross-modal retrieval} task: given a set of test-time ego-view images, we (i) compute their visual embedding and then (ii) retrieve a graph with the closest graph embedding in terms of cosine similarity. Given recent multi-city autonomous vehicle datasets~\cite{chang2019argoverse}, it is straightforward to construct pairs of ego-view images and street maps, both for training and testing. 

We train image and graph encoders to operate in the same embedding space, making use of recent techniques for cross-modal contrastive learning~\cite{radford2021learning}. Our key technical contribution is a novel but simple graph {\em encoder}, based on sequential transformers from the language community (\textit{i.e.}, BERT~\cite{devlin2018bert}) that extract fixed-dimensional embeddings from street maps of arbitrary size and topology. 

In fact, we find that even naive nearest-neighbor retrieval performs comparably to leading techniques for map generation~\cite{can2021structured,can2022topology}, \textit{i.e.}, returning the graph paired with the best-matching image in the training set. Moreoever, we demonstrate that cross-modal retrieval via \textit{Pix2Map} performs even better due to its ability to learn graph embeddings that regularize the output space of graphs. In addition, cross-modal retrieval has the added benefit of allowing one to expand the retrieval graph library with {\em un}paired graphs that lack camera data, leveraging the insight that retrieval need not be limited to the same (image, graph) training pairs used for learning the encoders. 
This suggests that \textit{Pix2Map} can be further improved with augmented road graph topologies that capture {\em potential} road graph updates (for which paired visual data might not yet exist). Beyond mapping, we show pilot experiments for visual localization, and the \textit{inverse} method, \textit{Map2Pix}, which retrieves a close-matching image from an image library given a graph. While not our primary focus, such approaches may be useful for generating photorealistic simulated worlds~\cite{mi2021hdmapgen}.

We summarize our \textbf{main contributions} as follows: We (i) show that dynamic street map construction from cameras can be posed as a cross-modal retrieval task and propose an contrastive image-graph model based on this framing. Building on recent advances in multimodal representation learning, we train a graph encoder and an image encoder with a shared latent space. 
We (ii) demonstrate empirically that this approach is effective and perform ablation studies to highlight the impacts of architectural decisions. Our approach outperforms existing graph generation methods from image cues by a large margin. 
We (iii) further show that it is possible to retrieve similar graphs to those in previously unseen areas without access to the ground truth graphs for those areas, and demonstrate the generalization ability to novel observations.

\section{Related Work}
\label{sec:relatedwork}

Maps are ubiquitous in robotics: given a pre-built map of the environment, autonomous agents can localize themselves via live sensory data and plan their future trajectories~\cite{deo2022multimodal}. 
Since the dawn of robotics, mapping and localization have been vibrant fields of research~\cite{Thrun05}, tackled using different types of sensors, ranging from line-laser RGB-D sensors for indoor mapping~\cite{rosu2020semi,cartillier2020semantic,grinvald2019volumetric}, to lidar~\cite{behley18rss} and/or cameras~\cite{Engel14ECCV,Engel15IROS,Muratal15TRO}, commonly used outdoors~\cite{Geiger12CVPR,caesar2020nuscenes,chang2019argoverse,sun20CVPR}. In the following, we focus on map construction and maintenance. For localization, we refer to prior work~\cite{Thrun05,ma2019exploiting}. 

\PAR{Map Representation.} Several map representations have been proposed in the community, ranging from full 3D maps, represented as meshes~\cite{rosu2020semi,Valentin13CVPR}, voxel grids~\cite{cartillier2020semantic,grinvald2019volumetric,Kochanov16IROS,Vineet15ICRA}, and (semantic) point clouds~\cite{cho2020semantic,behley18rss}. In visual localization~\cite{Sattler2011ICCV}, point clouds are often constructed using structure-from-motion methods \cite{Sattler2011ICCV} and additionally store visual descriptors that aid matching-based visual localization. The aforementioned representations can be used for highly accurate 6-DoF camera localization. However, they are heavy in storage (and, consequentially, transmission) \cite{zhou2022geometry}, which limits their applicability in outdoor environments.  

\PAR{High-Definition (HD) Maps.} Alternatively, High Definition (HD) maps store key semantic information, such as road layout and traffic light sign positions~\cite{Thrun2003RoboticMA}, together with their attributes and connectivity information. As shown in~\cite{ma2019exploiting}, such sparse and storage-efficient maps can be used as priors for centimeter-precise vehicle location in conjunction with vehicle sensors, such as cameras and lidars. 
While immensely useful, HD maps are difficult to create and maintain~\cite{liu2020high,seif2016autonomous, homayounfar2018hierarchical} and often require manual annotations and post-processing, rendering map construction and maintenance costly. Therefore, a problem of great importance is the automation of map construction and maintenance directly from sensory data. 

\PAR{HD Map Construction and Maintenance.} Several methods for HD map estimation rely on various sensor modalities. 
Li~\etal~\cite{li2019topological} propose a method that generates a topological map (represented as a spatial graph) of a city from satellite images. Wang \etal~\cite{wang2016torontocity} propose a collaborative approach that fuses several sources of information (data from airplanes, drones, and cars), such that consequent manual human post-processing can be minimized. Several methods tackle map construction directly from on-board vehicle sensory data. Often, methods tackle this challenging problem by first detecting road features in images (\eg, segment lanes)~\cite{behrendt2019unsupervised,det,pan2018SCNN,liu2021condlanenet,wu2021yolop}, and then utilize the camera and lidar sensory data to estimate precise road layout in 3D space. To generate spatial graphs, the aforementioned methods employ generative recurrent neural networks~\cite{liang2019convolutional,homayounfar2019dagmapper}, or optimization-based approaches~\cite{liebner2019crowdsourced}. Unlike the aforementioned works, we estimate spatial graphs directly from images, by-passing explicit lane estimation.

\begin{figure*}[t!]
\centering
   \includegraphics[width=0.94\linewidth]{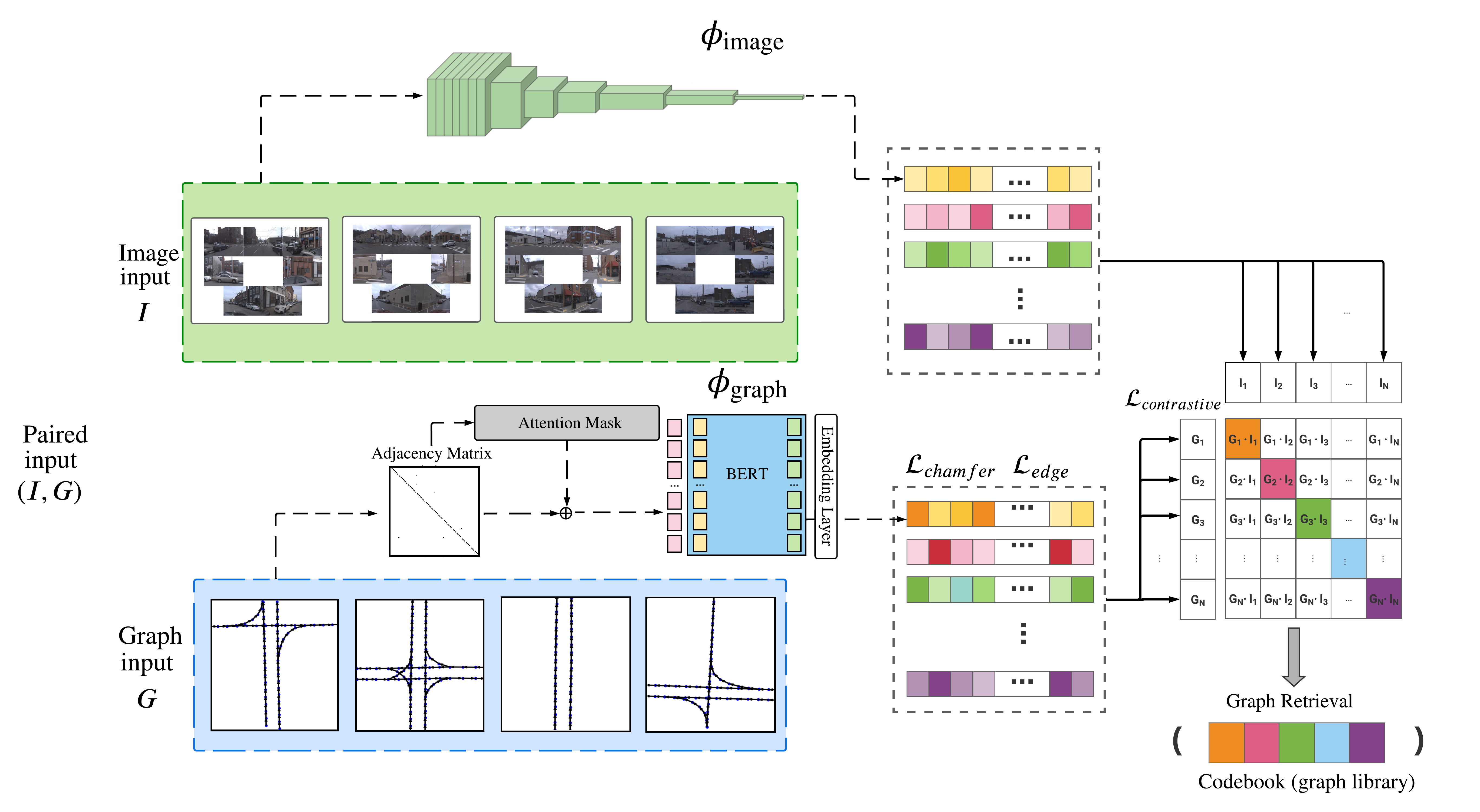}
     \vspace{-12pt}
   \caption{\textbf{Pix2Map}: The \textit{graph encoder (bottom)} computes a graph embedding vector $\phi_\text{graph}$ for each street map in a batch. The \textit{image encoder, (top)} outputs an image embedding $\phi_\text{image}$ for each corresponding image stack. We then build a similarity matrix for a batch, that contrasts the image and graph embeddings. We highlight that the adjacency matrix of a given graph is used as the attention mask for our transformer-based graph encoder.}
\label{fig:framework}
\end{figure*}

\PAR{Pixel Segmentation.} State-of-the-art methods for graph generation from monocular images~\cite{can2021structured,can2022topology} tend to first segment road lanes or centerlines in images, followed by graph generation using Polygon-RNN~\cite{castrejon2017annotating}. Instead of road lanes, HDMapNet~\cite{li2021hdmapnet} utilize methods for semantic/instance BEV maps (from cameras and/or lidar \eg, \cite{roddick2020predicting,yang2021projecting,saha2022translating, hu2021fiery, gilles2022gohome}), followed by heuristic post-processing to obtain vectorized HD maps. HDMapGen~\cite{mi2021hdmapgen} builds on recent developments in generative graph modeling~\cite{you2018graphrnn} to construct control points of central lane lines and their connectivity in a hierarchical manner. However, generating graphs \textit{conditioned} on a particular (image) input remains an open problem. Unlike HDMapGen, \textit{Pix2Map} sidesteps generative modeling, and instead directly retrieves a graph from a large database whose embedding vector is \textit{most similar} to image embeddings in terms of cosine distance. We show that \textit{Pix2Map} can also be used to keep HD maps up-to-date~\cite{lambert2021trust,pannen2019hd,bhavsar2020sensor,berrio2021long}.

\section{Method}
\label{sec:method}
In this section, we formalize our \textit{Pix2Map} approach as a cross-modal retrieval task. As shown in Fig.~\ref{fig:framework}, given training pairs of images and graphs, we learn image and graph encoders that map both inputs to a common fixed-dimensional space via contrastive learning. We then use the learned encoders to retrieve a graph (from a training library) with the most similar embedding to the test image.

\subsection{Problem Formulation}
\label{sec:problemformulation}
\begin{figure}[ht]
\centering
  \includegraphics[width=\linewidth]{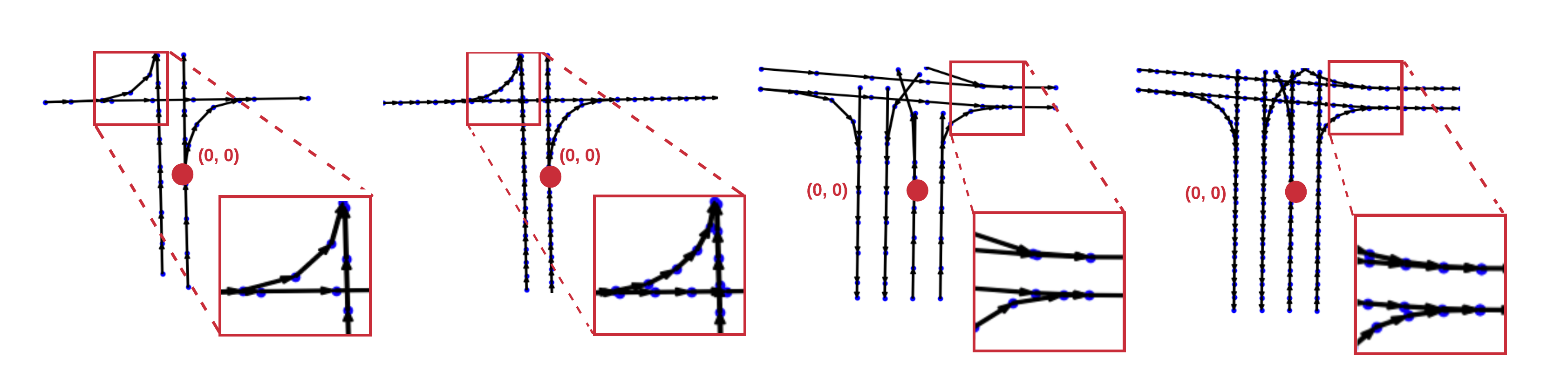}
  \vspace{-20pt}
  \caption{Two street map examples from Pittsburgh (\textit{left}) and Miami (\textit{right}) as \textit{segment graphs} and our resampled \textit{node graph}, with ego-vehicle origin at $(0,0)$.}
\label{fig:origin}
\end{figure}

We construct a library of image-graph pairs $(I, G)$ where $I$ is a list of $7$ ego-view images from a camera ring and $G$ is a street map represented as a graph $G = (V, E)$. Each vertex $v \in V$ represents a lane {\em node} and $E \in \{0,1\}^{|V| \times |V|}$ that encodes the connectivity between nodes, stored in an adjacency matrix. Lane nodes have a position attribute $(x,y)$ in a local egocentric ``birds-eye-view" coordinate frame, such that $(0,0)$ is the ego-vehicle location (Fig.~\ref{fig:origin}). Importantly, different graphs $G$ may have different numbers of lane nodes and connectivity information. 

\PAR{\bf Graph Representation.} The Argoverse dataset represents a street map as a \textit{segment graph}, where each vertex represents a lane {\em segment}. Lane segments are represented as polylines with $10$ $(x,y)$ points. We convert this \textit{segment graph} to a \textit{node graph} by defining each $(x,y)$ point as a graph node and adding a directed edge between successive points in a polyline (Fig.~\ref{fig:origin}). We further resample the \textit{segment graphs} by fitting degree-3 spline curves to lane segments, ensuring that connected nodes throughout the graph are approximately equidistant ($2m$).
We use this library for training the image and graph encoders, as detailed below.

\subsection{Image Encoder} 
\label{sec:imageencoder}
Given an image $I$, we use ResNet18 \cite{he2016deep} as a feature extractor (without fully connected layers) as an image encoder that learns fixed-dimensional embedding vectors $\phi_\text{image}(I) \in R^{512}$. 
To process $n$ input images (we use $n=7$ images throughout our work), we stack them channel-wise. 
We experiment with ImageNet-pre-trained weights, as well as training ``from scratch". In the pre-trained case, we replace the first convolution layer with one that stacks the original pre-trained filters $n$ times, with each weight divided by $n$. We reuse the original weights when making the new convolutional filters so the benefits of the pre-trained weights will be preserved. We ablate different training strategies and encoder architectures in Sec.~\ref{sec:experiments}.

\subsection{Graph Encoder} 
\label{sec:graphencoder}
Given a graph $G =(V, E)$, we would like to produce a fixed dimensional embedding $\phi_\text{graph}(G) \in R^{512}$, which is invariant to the orderings of the graph nodes.
Unlike pixels in an image or words in a sentence, nodes in graphs do not have an inherent order. 
We construct such a graph encoder using a Transformer architecture inspired by sequence-to-sequence architectures from the language \cite{devlin2018bert} defined on sequential tokens. Our encoder treats lane nodes as a collection of tokens and edges as masks for attention processing:
\begin{equation}
    \!\!\!\!\!\!v_{l+1} = \!\!\!\!\!\!\sum_{\{w: E(v,w) = 1\}} \!\!\!\!\!\!\!\!\!\text{Value}(v_l)\text{Softmax}_w [\text{Query}(v_l)\text{Key}(w)], \!\!\!\!\!\! %\nonumber 
\end{equation}
where $v_l$ is the embedding for vertex $v$ at layer $l$ and $v_0$ is initialized to its position $(x,y)$. We omit multiple attention heads and layer norm operations for brevity.

Embeddings are fed into a Transformer that computes new embeddings by taking an attention-weighted average of embeddings from nodes $w$ adjacent to $v$ (as encoded in the adjacency matrix $E$). 
We apply $M=7$ transformer layers (similar to BERT~\cite{devlin2018bert}). 
Finally, we average (or mean pool) all output embeddings to produce a final fixed-dimensional embedding for graph $G$, regardless of the number of nodes or their connectivity: $\phi_\text{graph}(G) = \frac{1}{|V|}\sum_{v \in V} v_M \in R^{512}$. 
In Sec.~\ref{sec:experiments} we ablate various design choices for our graph encoder, including the usage of periodic positional embeddings and encoding the edge connectivity information.

\subsection{Image-Graph Contrastive Learning}
\label{sec:loss}
To learn a joint embedding space, we follow the cross-modal contrastive formalism of~\cite{radford2021learning}, and briefly describe it here for completeness. Given N image-graph pairs $(I, G)$ within a batch, our model jointly learns the encoders $\phi_\text{image}(\cdot)$ and $\phi_\text{graph}(\cdot)$ such that the cosine similarity of the $N$ correct image-graph pairs will be high and the $N^2-N$ incorrect pairs will be low. We define cosine similarity between image $i$ and graph $j$ as: 
\begin{equation}
    \alpha_{ij} = \frac{\langle\phi_\text{image}(I_i) , \phi_\text{graph}(G_j)\rangle}{||\phi_\text{image}(I_i)||  ||\phi_\text{graph}(G_j)||}.
\end{equation} 
We then compute bidirectional contrastive losses composed of an image-to-graph loss $\ell^{(I \rightarrow G)}$ and a graph-to-image loss $\ell^{(G \rightarrow I)}$, following the form of the InfoNCE loss~\cite{oord2018representation}:
\begin{align}
    \ell^{(I \rightarrow G)}_{i} &= - \log \frac{\exp \alpha_{ii}}{ \sum_j \exp \alpha_{ij}},\\
    \ell^{(G \rightarrow I)}_{i} &= - \log \frac{\exp \alpha_{ii}}{ \sum_j \exp \alpha_{ji}}.
\end{align}
The contrastive loss is then computed as a weighted combination of the two, averaged over all positive image-graph pairs in each minibatch: 
\begin{equation}
    \ell_{contrastive} = \frac{1}{2N} \sum^{N}_{i=1} \Big( \ell^{(I \rightarrow G)}_{i} + \ell^{(G \rightarrow I)}_{i} \Big). \label{eq:clip}
\end{equation}
The above penalizes all incorrect image-graph pairs equally. We found it beneficial to penalize false matches between pairs with similar graphs (measured using graph metrics, Sec.~\ref{sec:experiments}) less severely, as similar graphs should intuitively have similar embeddings. We measure this similarity of the graphs after aligning the vertices. Formally, given a ground truth graph, $G_0$ and candidate match $G_i$, we first establish a correspondence between each vertex $v \in V_0$ and its closest match $\pi_i(v) = v_i \in V_i$ (in terms of Euclidean distance between vertices). Given such corresponding vertices, we compute both a Chamfer Distance~\cite{barrow1977parametric} (CD) and a binary cross-entropy (BCE) loss between the ground-truth binary adjacency matrix $E_0$ and the permuted matrix $E_i$: 
\begin{equation}
     \ell_{\mathit{chamfer}} = \sum_{v \in V_0} \sum_i \alpha_{i} \text{Distance}(v,\pi_i(v)), \label{eq:edgeloss}
\end{equation}
\begin{equation}
\!\!\!\!\!\!\!\ell_{edge} = \!\!\!\!\!\sum_{v,w \in V_0} \!\!\!\text{BCE} (\sum_i \alpha_i E_i( \pi_i(v),\pi_i(w)) %+ \frac{1 - \alpha_i}{2}
+\epsilon, E_0(v, w) ),\!\! %\nonumber
\end{equation}
where $\alpha_i = \softmax_i \alpha_{i0}$. The final loss is then: 
\begin{equation}
\ell = \omega_1\ell_{contrastive} + \omega_2\ell_{\mathit{chamfer}} + \omega_3\ell_{edge},
\end{equation} 
where $\omega_1 = 1$, $\omega_2 = 1$, $\omega_3 = 1/10$. 
To ensure that the BCE loss remains finite, we add a small non-zero $\epsilon$ to ensure that the edge probabilities are strictly positive. 
To speed up the loss computation, we ignore edges $v,w \in V_0$ that are missing for all graphs in the batch $E_i(\pi_i(v),\pi_i(w)) = 0, \forall i$. 
A key difficulty in evaluating graph edge losses such as our own or the Rand Loss (described in Sec.~\ref{Metrics}) is that they assume that a vertex-wise correspondence is already known between the predicted and target graphs. 
A more theoretically optimal framework may {\em search} over one-to-one vertex correspondences that jointly minimize the Chamfer and edge loss, \textit{e.g.}, by solving a bipartite matching problem~\cite{karp1990optimal}.

\subsection{Pix2Map via Cross-Modal Retrieval}
Given the learned encodings above, we now use them for regressing maps from pixel image input via retrieval. Denoting a graph library as $\mathbb{G}$, we retrieve image $I \Rightarrow G^*$, where $G^* = \argmax_{G \in \mathbb{G}_{\text{retrieval}}} \langle \phi_\text{image}(I), \phi_\text{graph}(G) \rangle$.
Note that the graph library used for retrieval $\mathbb{G}_{\text{retrieval}}$ need not be the same as the one used to train the image-graph encoders. Formally, let encoders be trained on a collection of image-graph pairs, written as $\mathbb{D}_{\text{train}} 	\coloneqq \{(I,G) |  I \in \mathbb{I}_{\text{train}},G \in \mathbb{G}_{\text{train}} \}$. $\mathbb{G}_{\text{retrieval}}$ need not be equivalent to $\mathbb{G}_{\text{train}}$, and furthermore, the set of corresponding images $\mathbb{I}_{\text{retrieval}}$ is not needed. 

This has several important properties. \textit{Firstly}, we can populate the graph library $\mathbb{G}_{\text{retrieval}}$ with additional graphs, not available during training, or cull the library to a particular subset of training graphs corresponding to a given city neighborhood. 
\textit{Secondly}, we can enlarge the graph library with graphs that have no corresponding images, including augmented variants of real street maps that capture potential map updates (such as the potential addition of a lane at a particular intersection, for which no real-world imagery would be available). 
\textit{Thirdly}, the above algorithm returns a ranked list of graphs, including near-ties. This can be used to generate multiple graphs that could correspond to an image input. 
\textit{Finally}, similar observations hold for retrieving street map images using a graph (i.e., \textit{Map2Pix}). \textit{Map2Pix} is more likely to be a one-to-many task, as the same street geometry can be associated with different visual pixels depending on the time of day or weather conditions.

\section{Experiments}
\label{sec:experiments}

In this section, we first discuss our evaluation test-bed (Sec.~\ref{sec:exp:testbed}) that we use to conduct the experimental evaluation. Then, we perform ablation studies to highlight each component's contribution (Sec.~\ref{sec:exp:ablations}). We compare our method to a recent state-of-the-art in Sec.~\ref{sec:exp:bench} and, finally, highlight several use cases of our \textit{Pix2Map} to automated map maintenance and expansion, and vehicle localization. 

\subsection{Evaluation Test-Bed}
\label{sec:exp:testbed}

\PAR{Dataset.} For evaluation we use Argoverse dataset~\cite{chang2019argoverse}, which provides seven ring camera images ($1920 \times 1200$) recorded at 30 Hz with overlapping fields of view, providing 360$^{\circ}$ coverage. Crucially, Argoverse contains street maps that capture the geometry and connectivity of road lanes. Such map annotations are not available in other autonomous vehicle datasets such as nuScenes~\cite{caesar2020nuscenes}. 
We perform the experiments across two cities in the United States, including Pittsburgh ($86km$) and Miami ($204km$). 
\PAR{Splits.} Argoverse provides train, validation, and test splits. Note that validation and test splits may include regions that spatially overlap with the regions included in the training set. However, recordings of these regions were collected at different data collection runs at different times. To evaluate realistic applications of \textit{map-updating} (where one trains on, \eg, Pittsburgh up to 2021 and tests on Pittsburgh 2022+) and \textit{map-expansion} (where one trains on the neighborhood of Squirrel Hill and tests on Shadyside), we split up the union of (test+val) into those regions that spatially overlap the trainset and those that do not. We refer to these as \textit{MapUpdate} and \textit{MapExpand} test sets (Fig.~\ref{fig:dataset}). We present results for both settings, but default to \textit{MapUpdate} for diagnostics unless otherwise specified.

\PAR{\bf Map Preprocessing.} The key component of HD maps is the central line of drivable lanes. We extract the subgraphs corresponding to $40m \times 40m$ spatial windows. 
We use the adjacency matrix to represent the node connectivity. 
An edge connects two nodes if they are immediately reachable following the traffic flow \ie, a subgraph of nodes in a given lane corresponds to a directed path. 
Moreover, an edge exists between two lanes if the first node of the second lane follows directly from the last node of the first, either because one lane continues to another or because one can turn from one lane to the other. 
We make sure to rotate the node positions and lanes to align with the driving direction. 

\PAR{\bf Implementation.} We train on a single NVIDIA A100 GPU, and the training dataset contains up to 512 samples in one batch. The model is trained for a total of 40 epochs, where a single epoch takes 40 minutes of wall-clock time. We use the Adam optimizer with a learning rate of $2\text{e-}4$. We use a pretrained ResNet18 for the image encoder. In order to support an input containing several images, we duplicate and stack the filters of the input conv layer corresponding to the number of images. We then divide the parameters by the number of images per input example, giving us a model that initially returns an identical output to the original model. We extract the feature representation immediately before the fully connected layer.
For the graph encoder, we use a Bert model with mean pooling and no positional embeddings to model the pairwise intersections between each of the nodes. For each node, we pass its adjacencies and its coordinates. We apply the attention mask which indicates to the model which tokens should be attended to and which should not.

\subsection{Metrics}
\label{Metrics}
To quantitatively evaluate the quality of the retrieved graphs, we design three types of metrics to capture the difference between the retrieved graph $G_1 = (V_1, E_1)$ and the ground truth $G_2 = (V_2, E_2)$.

\noindent{\bf Spatial Point Discrepancy.} We first introduce metrics that represent lane graph nodes $v \in V$ as $(x,y)$ points corresponding to the lane centroid, ignoring edge connectivity. We can then use metrics to measure differences between set of points. {\em Chamfer Distance} computes the closest point in $v_2 \in V_2$ for every $v_1 \in V_1$ (and vice versa, to ensure symmetry). 
{\em Maximum Mean Discrepancy (MMD)}~\cite{hajiramezanali2019variational,mi2021hdmapgen} measures the squared distance between point centroids in a Hilbert space using Gaussian kernels $\langle \varphi(v_1),\varphi(v_2)\rangle_H = k(x_1-x_2,y_1-y_2)$:
\begin{align}
&\text{MMD}(G_1,G_2) = \left\Vert\frac{1}{|V_1|} \sum_{v_1 \in V_1} \varphi(v_1) - \frac{1}{|V_2|}\sum_{v_2 \in V_2} \varphi(v_2)\right\Vert^2_H \label{eq:mmd}. %\\
%& = \frac{1}{n^2}\sum_{ij} \Big( k(x_i, x_j) + k(y_i, y_j) - 2k(x_i, y_j) \Big)
\nonumber
\end{align}    

\noindent{\bf Edge Connectivity.} The above metrics evaluate the quality of only the retrieved graph nodes, but not their edge connectivity. We define a RandLoss similar to \eqref{eq:edgeloss}, as: 
\begin{align}
\text{RandLoss} = \sum_{v,w \in V_1} \mathds{1}_{[E_2(\pi(v),\pi(w)) \neq E_1(v, w)]},
\nonumber
\end{align}
where $\mathds{1}$ is an indicator function for mismatching edge labels between a pair of nodes in the graph $G_1$ and their corresponding pair in graph $G_2$. 
This metric is also known as the Rand index for clustering evaluation~\cite{rand1971objective}. 

\noindent{\bf Urban Planning.} We also report a set of metrics motivated by the urban planning literature~\cite{mi2021hdmapgen,chu2019neural,alhalawani2014makes}, evaluating the degree to which we are able to reconstruct the following key properties of urban HD maps. {\bf Connectivity} is the number of edges relative to the number of lane nodes. {\bf Density} is the number of edges relative to the maximum possible number of edges. {\bf Reach} is designed to capture urban development and is defined as the total distance covered by the lanes:
\begin{align}
    &\text{Connectivity} = \frac{\|E\|_0}{|V|}, \quad 
    \text{Density} = \frac{\|E\|_0}{|V|(|V| - 1)}, \nonumber \\
    &\text{Reach} = \sum_{(v,w): E(v,w)=1}\text{len}(v,w).
\nonumber
\end{align} 
We report the absolute relative error~\cite{mi2021hdmapgen} for these metrics.

\begin{table}[t!] 
\centering
\resizebox{1.0\linewidth}{!}{
\begin{tabular}{c| c |  c | c c c c c c c}
\toprule
Methods  &	Chamfer & RandLoss &	MMD	&	U. density	&	U. reach	&	U. conn.		\\
 & $10^{1}$ & $10^{-2}$ & $10^{-1}$ & $10^{-1}$ & $10^{-1}$ & $10^{-1}$  \\
\midrule
PINET \cite{ko2021key} & 4.9244 &10.8935&4.2983&2.8194&7.4194& 2.9231\\
TOPO-PRNN \cite{can2022topology} & 7.4811 & 9.2813 &  5.7726 & 3.9371 & 6.8297 & 1.3934	\\
TOPO-TR \cite{can2022topology}  & 3.0140 &\textbf{7.1603} &4.6431 & 2.2467 & 3.3091 & 1.1530\\
\textit{Pix2Map}-Unimodal & 4.3967 & 9.0764 & 4.1873 & 1.8391 & 3.2746 & 1.7734 \\
\textit{Pix2Map}-Single & 2.6819 & 7.5204 & 4.0848 & 2.5339 & \textbf{3.0134} &  \textbf{1.0291}\\
\textit{Pix2Map} (ours) & \textbf{2.0882} & 7.7562 & \textbf{3.9621} & \textbf{1.4354} & 3.2893 &1.5532	\\
\bottomrule
\end{tabular}
}
\vspace{-7pt}
\caption{\textbf{Baseline comparisons.} For fair comparisons with the prior art~\cite{can2022topology}, in this experiment, we (i) train \textit{Pix2Map} using frontal $50m\times50m$ road-graphs (as opposed to our default setting of predicting the surrounding $40m\times40m$ area).
Moreover, we (ii) train \textit{Pix2Map} with a single frontal view (\textit{Pix2Map}-Single) to ensure
consistent comparisons to baselines. 
Importantly, even in this setting, \textbf{our method still outperforms baselines by a large margin}: $2.6819$ in terms of Chamfer distance, as compared to $3.0140$ obtained by the closest competitor, TOPO-TR~\cite{can2022topology}.} 
\label{tab:bench_ours}
\end{table}

\subsection{Baselines}
\label{sec:exp:bench}

\begin{figure}[t!]	
\centering	
  \includegraphics[width=\linewidth]{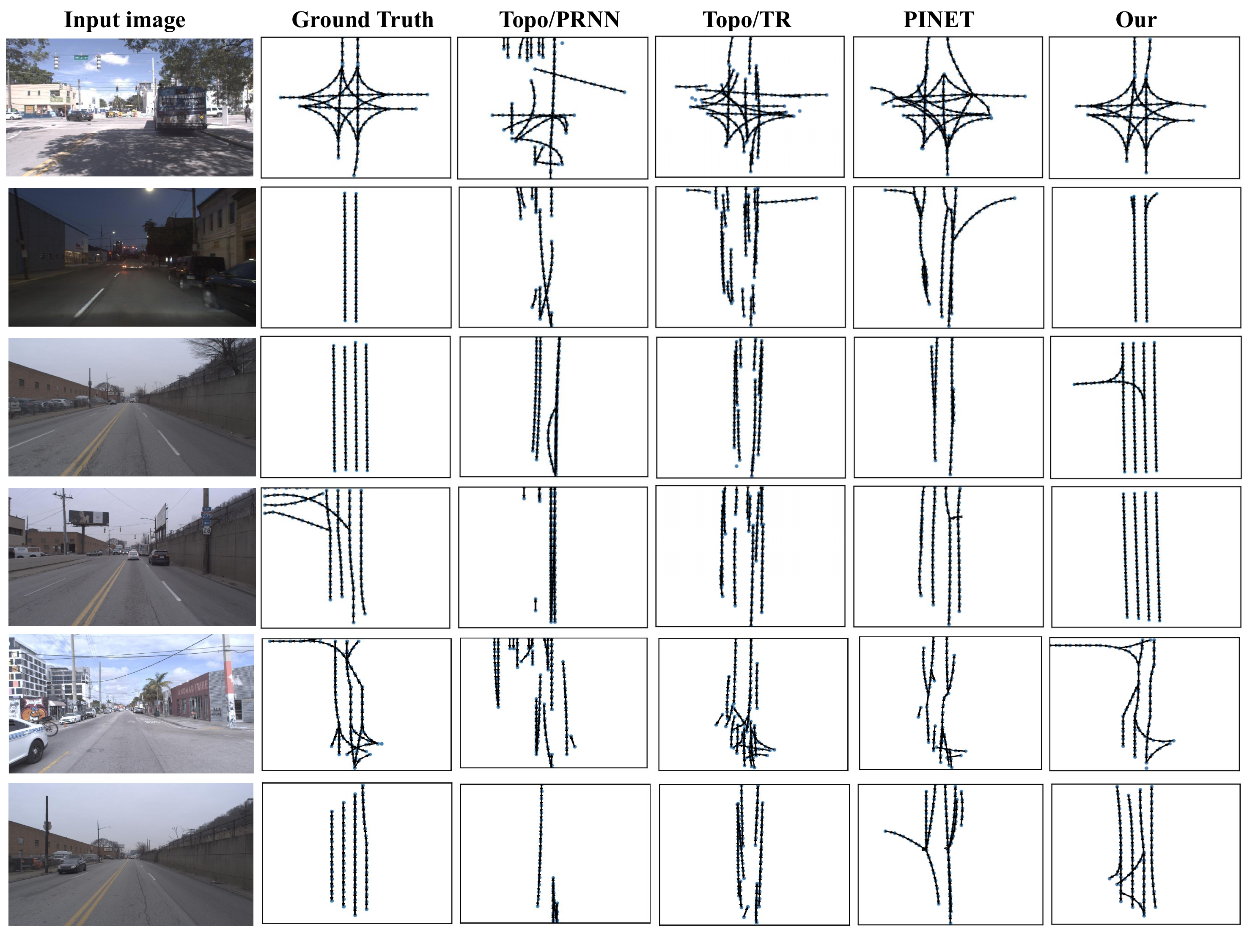}
  \vspace{-25pt}
  \caption{{\bf Qualitative results}. From left to right: input image, ground-truth maps, maps generated by state-of-the-art methods, and, in the last column, our method. As can be seen, the retrieved maps with our method have the highest visual fidelity. 
  }
    \label{tab:qualitative_baselines}
\end{figure}

\begin{figure*}[ht]
\centering
 \includegraphics[width=0.85\linewidth]{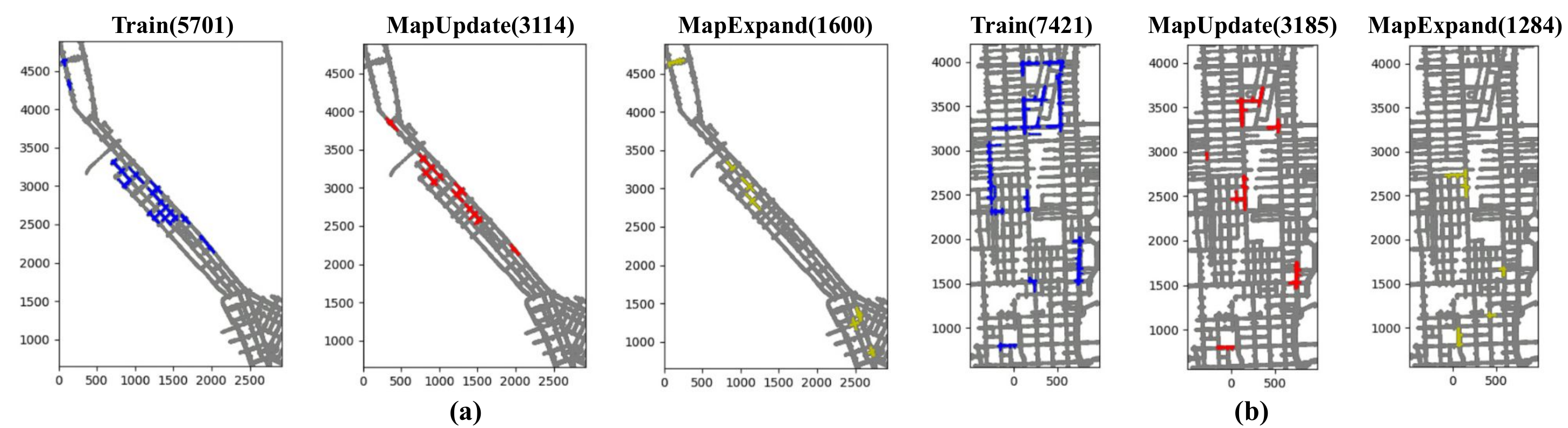} 
 \vspace{-10pt}
   \caption{\textbf{Pittsburgh (a) and Miami (b) datasets.} Including training data (blue), MapUpdate test data (red) that overlap the blue but are collected at different times, and MapExpand test data (yellow) that do not non-overlap. We denote the size of each dataset (in terms of the number of image-graph pairs) in parentheses.}
\label{fig:dataset}
\end{figure*}

We show a visual comparison with top performers in Fig.~\ref{tab:qualitative_baselines} and quantitative results in Tab.~\ref{tab:bench_ours}. 
We first report the performance of a naive nearest-neighbor baseline (Unimodal), which returns the graph associated with the closest training image example. This unimodal approach already performs on par with the state-of-the-art Transformer and Polygon-RNN~\cite{homayounfar2018hierarchical} based methods TOPO~\cite{can2021structured}, and PINET~\cite{ko2021key}. 
Our diagnostics further explore the improvement from unimodal to cross-modal retrieval. As can be seen, \textit{Pix2Map} improves greatly over several state-of-the-art baselines in terms of the Chamfer distance, MMD, urban density error, and urban connectivity error, while performing comparably in terms of RandLoss and urban reach error. Moreover, our method is especially strong in terms of preserving the spatial point discrepancy, outperforming baselines by a large margin. We note that~\textit{Pix2Map} was designed to fully utilize the image data available in the camera ring, whereas baselines use only frontal view. For apples-to-apples comparison, we retrain \textit{Pix2Map}
with a single frontal view (\ie, single camera, \textit{Pix2Map}-Single. As can be seen, even the single-view variant of \textit{Pix2Map} outperforms the closest competitor (TOPO-TR) across almost all metrics except RandLoss and Urban density. Our results suggest that baselines may
also benefit from multi-view processing.

% Put the table first since we talk about the setup

\begin{table*}[ht]
\centering
\resizebox{1.0\linewidth}{!}{
\begin{tabular}{c| c|  c |c | c |  c | c c c c c c}
\toprule
Row & 	$\mathcal{E}_{img}$  	&	Attention 	&  Adjacency  &	Positional 	& Resampling &	Chamfer 	& RandLoss &	MMD	&	U. density	&	U. reach	&	U. conn.	\\
  &   & Mask &  Matrix & Encoding &  & $10^{1}$ & $10^{-2}$ & $10^{-1}$ & $10^{-1}$ & $10^{-1}$ & $10^{-1}$ \\
\midrule
1	&	1 $\times$ RN18	&	\Checkmark	&		&		&	\Checkmark	&	1.9241	&	9.0446	&	4.1804	&	1.2058	&	3.6333	&	1.8398	\\														
2	&	1 $\times$ RN18	&		&	\Checkmark	&		&	\Checkmark	&	1.9309	&	8.3834	&	3.8383	&	1.1934	&	3.7428	&	1.8629	\\														
3	&	1 $\times$ RN18	&	\Checkmark	&	\Checkmark	&		&	\Checkmark	&	\textbf{1.5908}	&	7.3283	&	\textbf{3.0888}	&	\textbf{0.7593}	&	\textbf{3.2997}	&	\textbf{0.8397}	\\								
4	&	1 $\times$ RN18	&	\Checkmark	&	\Checkmark	&		&		&	3.2663	&	\textbf{6.9943}	&	6.3704	&	3.6883	&	5.3219	&	4.1658	\\														
5	&	1 $\times$ RN18	&	\Checkmark	&	\Checkmark	&	\Checkmark	&	\Checkmark	&	2.1564	&	9.1200	&	8.8328	&	0.8813	&	3.4290	&	1.5481	\\														
6	&	7 $\times$ RN18	&	\Checkmark	&	\Checkmark	&		&	\Checkmark	&	4.8129	&	11.2118	&	9.7538	&	3.9169	&	6.7794	&	2.5285	\\												\bottomrule
% \hline
\end{tabular}
}
\vspace{-7pt}
\caption{\textbf{Image and graph encoder ablations.} From left to right, we ablate a) encoding each of the seven ego-images separately or using an early-fusion multiview image encoder, b) restricting the transformer attention mask to the graph-adjacency matrix or using the default fully-connected attention,  c) the inclusion of the corresponding row of the attention matrix as a node input feature for the model, d) adding a positional encoding to each graph vertex, and e) resampling graph vertices to be equidistant. We find results are dramatically improved by early fusion for image encoding (row3-vs-row6) and graph vertex resampling (row3-vs-row4). Results are marginally improved by restricting the attention mask (row2-vs-row3) and adding the adjacency matrix as an input feature (row1-vs-row3 and row2-vs-row3). Perhaps surprisingly, adding in positional vertex encodings slightly decreases performance (row3-vs-row5). 
}
%\vspace{-7pt}
\label{tab:ablation_encoders}
\end{table*}

\begin{table}[ht]
\centering
%\vspace{-10pt}
%\footnotesize
\resizebox{1.0\linewidth}{!}{
\begin{tabular}{ c c c |c c c c c c}
\toprule
% \hline
  $\mathcal{L}_{con.}$  &  $\mathcal{L}_{edge}$ & $\mathcal{L}_{chamf.}$  & Chamfer & RandLoss & MMD  &  U. density & U. reach & U. conn. \\ 
 & &  & $10^{1}$ & $10^{-2}$ & $10^{-1}$ & $10^{-1}$ & $10^{-1}$ & $10^{-1}$  \\
\midrule
\checkmark &  &  & 
 3.2249	&	9.1727	&	7.1070	&	2.7387	&	14.1033	&	3.6998\\
 \checkmark & \checkmark &&  
2.7967	&	8.9951	&	8.8717	&	0.9808	&	10.0251	&	1.0512\\
 \checkmark & & \checkmark & 
 1.7440	&	8.5186	&	\textbf{2.2480}	&	0.9841	&	7.4534	&	1.5664\\
 \checkmark &\checkmark & \checkmark &  
\textbf{1.5908}	&	\textbf{7.3283}	&	3.0888	&	\textbf{0.7593}	&	\textbf{3.2997}	&	\textbf{0.8397}	\\
\bottomrule
% \hline
\end{tabular}
}
\vspace{-7pt}
\caption{\textbf{Ablation on the training loss.} Adding a partial credit for matching to graphs with low Chamfer distance ($\mathcal{L}_{cham.}$) to the ground truth improves results considerably compared to vanilla contrastive loss ($\mathcal{L}_{con.}$). By adding in an additional edge loss ($\mathcal{L}_{edge}$), we further improve the performance.}
\label{tab:ablation_loss} 
\end{table}

\begin{table}[t] %[t!]
\centering
%\vspace{-15pt}
%\vspace{-10pt}
%\footnotesize
\resizebox{1.0\linewidth}{!}{
\begin{tabular}{c|c|c|cccc}
\toprule
Methods  &	Chamfer & RandLoss &	MMD	&	U. density	&	U. reach &	U. conn. \\
& $10^{1}$ & $10^{-2}$ & $10^{-1}$ & $10^{-1}$ & $10^{-1}$ & $10^{-1}$ \\
\midrule
Unimodal & 3.2168 & 9.7596 & 7.7671 & 0.7365 & 3.9452 & 1.3661 \\
Ours & 1.5908	&	7.3283	&	3.0888	&	0.7593	&	3.2997	&	0.8397	\\
Ours++   &1.5208	&	6.1504	&	3.0944	&	0.7407	&	3.2610	&	0.8089\\
\bottomrule
% \hline
\end{tabular}
}
\vspace{-7pt}
\caption{Cross-model retrieval ({\bf ours}) significantly outperforms classical {\bf unimodal} retrieval (\ie, the nearest neighbor on image encoder features). Cross-modal retrieval can exploit the graph embedding space, which appears to regularize retrieval results, while the unimodal approach does not utilize any graph embedding. Moreover, our cross-modal retrieval can take advantage of larger unpaired graph libraries, which further improve performance ({\bf ours++}). Unimodal retrieval requires paired data.} 
\label{tab:baseline_modalities}
\end{table}
% \vspace{-7pt}

\begin{table}[ht] %[t!]
\centering
%\vspace{-15pt}
%\vspace{-10pt}
%\footnotesize
\resizebox{1.0\linewidth}{!}{
\begin{tabular}{ c |  c | c c c c c c c}
\toprule
City & Library Size  &	Chamfer & RandLoss &	MMD	&	U. density	&	U. reach	&	U. conn.		\\
&  $10^{1}$ & $10^{-2}$ & $10^{-1}$ & $10^{-1}$ & $10^{-1}$ & $10^{-1}$  \\
\midrule
\marklg PIT& 5.7k & 1.5908	&	7.3283	&	3.0888	&	0.7593	&	3.2997	&	0.8397\\
\marklg &  10k  &1.6457	&	7.6247	&	3.2848	&	\textbf{0.7264}	&	4.5891	&	1.6364\\
\marklg & 20k  &1.5369	&	6.5373	&	3.1883	&	0.7581	&	3.2902	&	1.0602\\
\marklg & 30k  &1.5239	&	6.6553	&	\textbf{3.0253}	&	0.8586	&	4.0642	&	0.9615\\
\marklg & 40k  &\textbf{1.5208}	&	\textbf{6.1504}	&	3.0944	&	0.7407	&	\textbf{3.2610}	&	\textbf{0.8089}\\
\midrule
\marklgg MIA &7.4k & 1.4747	&	6.8693	&	3.4033	&	1.0948	&	4.6253	&	1.1910\\
\marklgg &10k & 1.4991	&	6.2315	&	3.3118	&	1.2784	&	5.5209	&	1.3679\\
\marklgg &20k & 1.3878	&	8.0234	&	3.3910	&	1.1290	&	4.1237	&	1.3249\\
\marklgg &30k & 1.4012	&	7.1898	&	3.2773	&	1.2444	&	4.2471	&	1.2298\\
\marklgg &40k & 1.3878	&	7.6305	&	3.3351	&	1.2523	&	5.3894	&	1.3385\\
\marklgg &60k & 1.3080	&	6.3369	&	3.18879	&	1.1972	&	4.7578	&	1.1977\\
\marklgg &80k & 1.2711	&	6.2852	&	3.19506	&	1.0123	&	4.6827	&	1.1651\\
\marklgg &100k & \textbf{1.2462}&\textbf{6.2740}& \textbf{3.1277}	&	\textbf{0.9884}	&	\textbf{3.8521}	&	\textbf{1.1397}\\
\bottomrule
\end{tabular}
}
\vspace{-7pt}
\caption{\textbf{Ablation with larger map-graph libraries.} As we grow the graph retrieval library (including maps without corresponding image views), we observe performance grows consistently with the size of the retrieval library. } % Map-expand}
\label{tab:ablation_enlarged}
\end{table}

\begin{figure*}[ht]
\centering  \includegraphics[width=\linewidth]{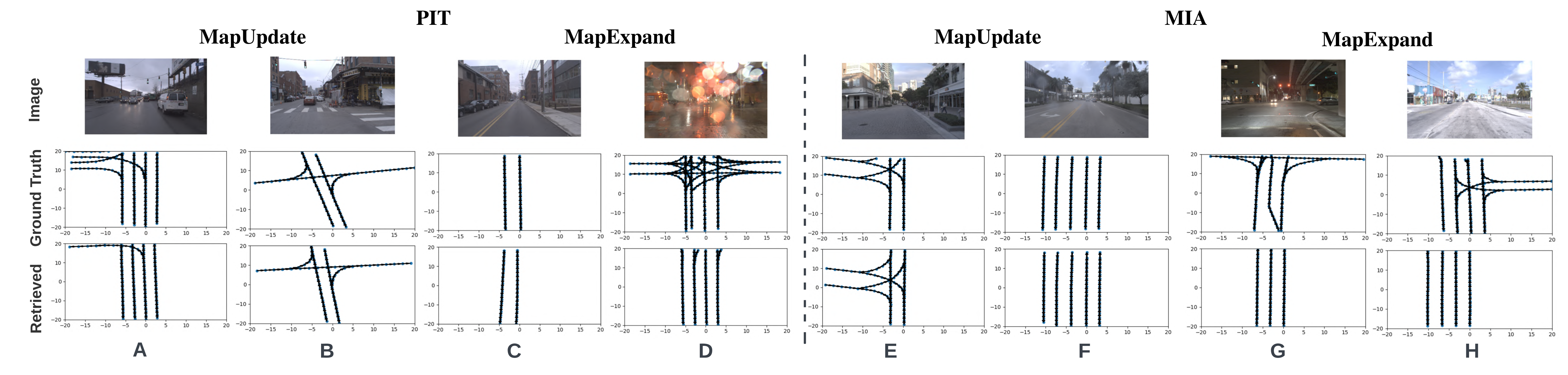}
  \vspace{-20pt}
  \caption{Given a set of ego-view images (front camera, \textit{top}), we plot the ground-truth graph (\textit{middle}), followed by the \textit{Pix2Map} predictions (\textit{bottom}) for both the \textbf{MapUpdate} (\textit{left}) and \textbf{MapExpand} (\textit{right}) tasks for two cities, Pittsburgh (\textbf{PIT}) and Miami (\textbf{MIA}). 
  In general, retrieval results are quite accurate, particularly for \textbf{MapUpdate}, where train and test samples are drawn from the same geographic regions. Inclement weather such as heavy rain is challenging (row \textbf{PIT MapExpand}, column \textbf{D}) due to the degraded visual signal. }
\label{fig:result}
\end{figure*}

\subsection{Model Analysis}
\label{sec:exp:ablations}

In this section, we experiment with two graph representations, evaluate different design decisions on the graph and image encoders, and multimodal contrastive learning.

\PAR{Ablations.} Tab.~\ref{tab:ablation_encoders} ablates several design decisions on image (Sec.~\ref{sec:imageencoder}) and graph (Sec.~\ref{sec:graphencoder}) encoders. For the analysis, we focus on Chamfer distance as an illustrative metric, as it appears to be most consistent with graph qualitative estimation. 
\PARit{Image Encoder.} As can be seen in Tab.~\ref{tab:ablation_encoders}, row3-vs-row6, \textit{early fusion} (1$\times$RN18) is significantly better compared to a \textit{late fusion} (7$\times$RN18) variant, where we separately encode images with a distinct ResNet18 model for each camera, followed by fusion via average-pooling. 

\PARit{Graph Encoder.}We consider three primary variations on the graph encoder architecture. The results suggest (Tab.~\ref{tab:ablation_encoders}, row2-vs-row3) that restricting the attention mask marginally improves the results. Furthermore, encoding the edge connectivity information (adding the adjacency matrix as an input feature, row1-vs-row3) slightly improves the performance. Perhaps surprisingly, positional vertex encodings slightly decrease performance (row3-vs-row5).

\PARit{Graph Representation.} Tab.~\ref{tab:ablation_encoders} (row3-vs-row4) shows that the proposed graph vertex resampling dramatically improves the results. As we mentioned in Sec.~\ref{sec:problemformulation}, we resample the \textit{segment graphs} and the connected nodes throughout the graph to be approximately equidistant. %, with a distance of 2 meters. 

\PARit{Loss.} Tab.~\ref{tab:ablation_loss} ablates our loss function (Sec.~\ref{sec:loss}). Compared to the na\"ive contrastive loss $\mathcal{L}_{contrastive}$, which weighs all incorrect image-graph matches equally, we find adding in partial credit for matching to graphs with low Chamfer distance $\mathcal{L}_{chamfer}$, to the ground-truth improves results considerably while adding in an additional edge loss $\mathcal{L}_{edge}$  further improves performance. 
\textit{We use the most performant combination (row3) for further experiments.}

\PAR{\bf Unimodal v.s. Cross-modal.} To quantify the benefits of the cross-modal training scheme, we compare \textit{Pix2Map} to its image encoder alone, evaluating it as a unimodal image-encoding-based retriever. Specifically, in \textit{Pix2Map}, we directly retrieve graphs by finding the graph embedding that is closest to the input image embedding in the multimodal embedding space. However, in this ablation, we instead find the image embedding in the training set which is closest to the input image embedding and return its corresponding graph. We find that using both modalities improves performance on almost all metrics, as shown in Tab.~\ref{tab:baseline_modalities}. The improvements range from a 37.0\% decrease in RandLoss to a 60.2\% decrease in MMD, with one increase of less than a percentage point in urban density error. However, note that while unimodal ablation performs broadly worse than \textit{Pix2Map}, it still performs better than any baseline shown in Tab.~\ref{tab:bench_ours} in terms of Chamfer, MMD, and Urban Density. 

\PAR{\bf Augmenting the Graph Library.} One of the benefits of our cross-modal retrieval approach is that we can match (or retrieve from) arbitrary collections of graphs that are different from (or larger than) the training graphs used to learn cross-modal encoders. This allows us to make use of graphs that do not have corresponding images. Interestingly, the Argoverse dataset provides such data, as maps include many locations for which no imagery is provided. By sampling random ego-vehicle positions in Miami and Pittsburgh, we grow both the Pittsburgh library and the Miami library to $40k$, thus significantly expanding our retrieval graph library. We summarize these results in Tab.~\ref{tab:ablation_enlarged}. We observe that performance improves across the suite of metrics as the graph retrieval library grows larger. This suggests there is a significant potential to further improve our results by simply growing our graph retrieval dataset using existing maps, without access to corresponding image pairs.

\PAR{Applications.} In the Appendix, we present several experiments that demonstrate the practical applications of \textit{Pix2Map}. In particular, we present experiments on expanding (\textit{MapExpand}) and updating (\textit{MapUpdate}) existing maps using \textit{Pix2Map}, see Fig.~\ref{fig:result}. Furthermore, we show that \textit{Pix2Map} can be used for visual localization by generating a heatmap of possible ego-vehicle locations in a city-level map based on input images. Finally, we visually demonstrate the inverse \textit{Map2Pix} task for retrieving ego-camera images given a street map. This experiment shows that it is possible to retrieve egocentric camera data using street maps, which can be used to synthesize virtual worlds consistent with the query road geometry.

\section{Conclusion}

In this work, we propose a significantly different approach to inferring high-definition maps from cameras. Rather than learning a nonlinear mapping from image pixels to BEV and generating a discrete spatial graph, we suggest a retrieval-based approach. Our experiments indicate that learning a multimodal embedding space for camera data and map data holds promise, and we hope our work can serve as an essential building block for map expansion and updating in the autonomous driving field. Beyond map maintenance, we also demonstrate that our approach can be used as a novel form of visual localization. While these results are encouraging, there are numerous potentially impactful future directions to explore. For instance, instead of performing graph retrieval, one could utilize the latent space to generate new unseen graphs using a graph-based decoder architecture.

%%%%%%%%% ACK
\footnotesize{\PAR{Acknowledgments.} This work was supported by the CMU Argo
AI Center for Autonomous Vehicle Research.

%%%%%%%%% REFERENCES
{\small
\bibliographystyle{ieee_fullname}
\bibliography{refs}
}

\clearpage

\setcounter{page}{1}

\twocolumn[
\centering
\Large
\textbf{Supplementary material for Pix2map: \\ Cross-modal Retrieval for Inferring Street Maps from Images} \\
\vspace{1.0em}
] 
\appendix

\normalsize

\begin{abstract}
In this supplement, we provide various experiments to illustrate the practical uses of \textit{Pix2Map}. These experiments include:
\begin{itemize}
\item Map Expansion and Update, in which we present experiments on expanding and updating existing maps,
\item Visual Localization, by generating a heatmap of possible locations for the ego-vehicle on a city-level map,
\item Map2Pix, which is visually demonstrated by retrieving ego-camera images using street maps. 
\end{itemize}

\end{abstract}

\section{Applications}
\label{sec:exp:applications}

In this section, we discuss how our method can be used for practical purposes, and show that graph library retrieval can greatly improve various downstream applications such as expansion (\textit{MapExpand}) and update (\textit{MapUpdate}) given existing maps,  visual image-to-HD map localization and \textit{Map2Pix}.

\subsection{Map Expansion and Update} 
\label{sec:mapupdate}
We use our graph retrieval method to mimic map expansion (\textit{MapExpand}) and map update (\textit{MapUpdate}) using data splits. 
For map expansion, we retrieve local graphs corresponding to recordings obtained in a ``new traversal'' to expand the existing map. For map updates, we similarly retrieve local maps to update the global map.

\begin{table}[h]
\centering
%\vspace{-10pt}
%\footnotesize
\resizebox{1.0\linewidth}{!}{
\begin{tabular}{ cc | c c c c c c}
\toprule
% \hline
 City & Task type  &  Chamfer & RandLoss	&	MMD	&	U. density	&	U. reach	&	U. conn.	\\
    &  & $10^{1}$ & $10^{-2}$ & $10^{-1}$ & $10^{-1}$ & $10^{-1}$ & $10^{-1}$   \\
\midrule
 \markg PIT & \markg MapUpdate & 
\markg 1.5908	&\markg 7.3283	&\markg 3.0888	&\markg 	0.7593	&\markg 	3.2997	&\markg 	0.8397	\\
 \markg  & \markg MapExpand & 
\markg 2.6654	&\markg 	16.9768	&\markg 	8.0468	&\markg 	3.9482	&\markg 	4.2949	&\markg 	3.9699\\
\midrule
 \markgg MIA & \markgg MapUpdate  &  \markgg 1.4747	&\markgg 	6.8693	&\markgg 	3.4033	&\markgg 	1.0948	&\markgg 	5.5333	&\markgg 	1.1910\\
 \markgg  & \markgg MapExpand  &\markgg 2.0637	&\markgg 	11.1354	&\markgg 	4.2605	&\markgg 	1.4922	&\markgg 	4.7318	&\markgg 	1.5940\\
\bottomrule
% \hline
\end{tabular}
}
\caption{\textbf{Map update and expansion evaluation.} As can be seen, map expansion to novel areas can be much harder than updating previously-seen areas.}
\label{tab:update_exp_eval}
\end{table} 

\begin{figure}[t]	
\centering	
   \includegraphics[width=\linewidth]{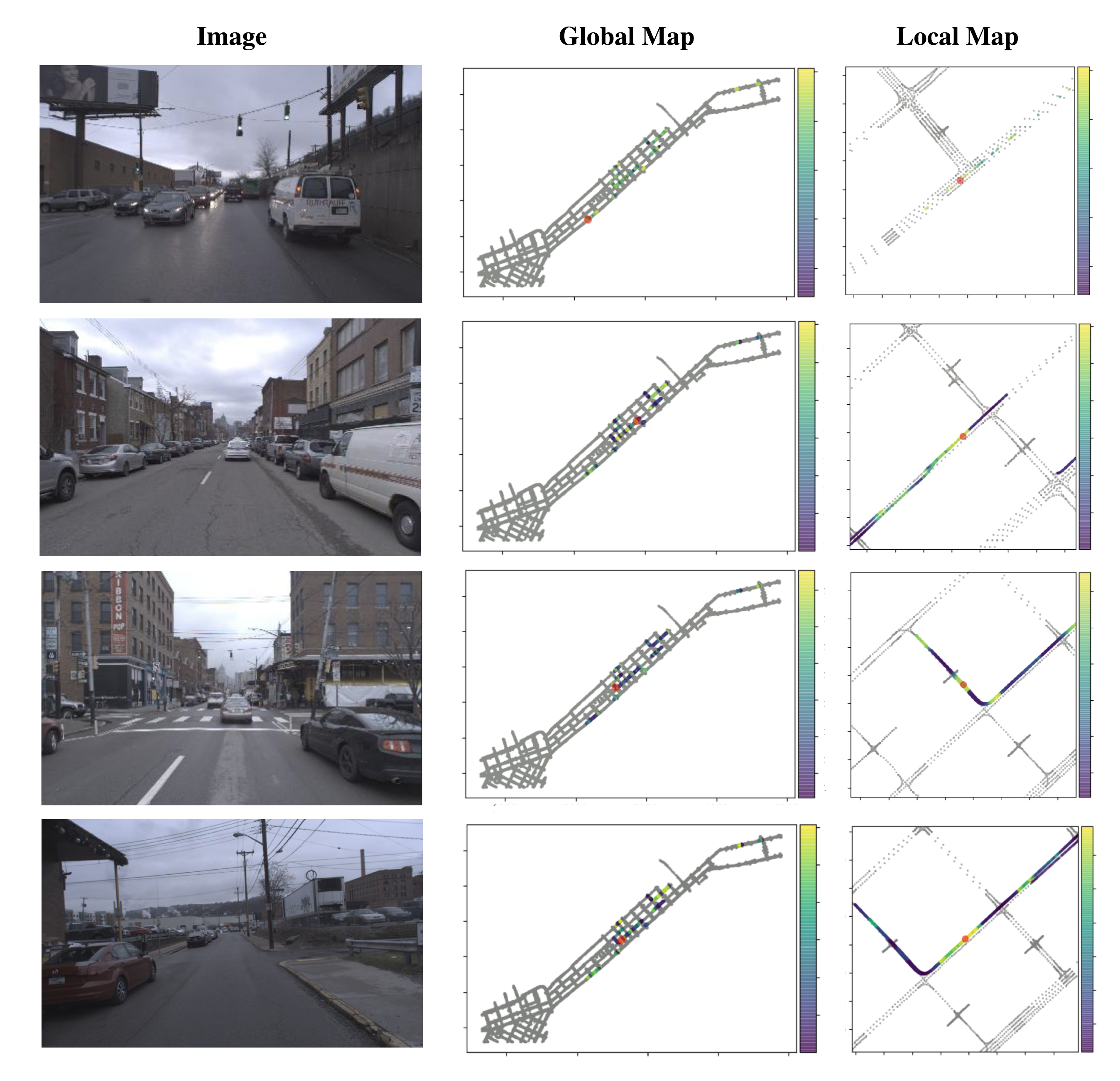}	
   \caption{{\bf Visual localization via Pix2Map.} We overlay retrieval scores on the corresponding local graphs from the original city map, generating a graph ``heatmap" of possible locations given instantaneous ego-view images. We plot the ground-truth location as a red dot. In general, ground-truth locations tend to lie in high-scoring (yellow) regions. For example, the top ground truth corresponds to an intersection, while other high-scoring regions also tend to be graph intersections as well. Given a sequence of images, one may be able to reduce the ambiguity over time~\cite{brubaker2015map}}	
\label{fig:localization_heatmap}	
\end{figure}

\begin{figure}[h!]
\centering
  \includegraphics[width=0.9\linewidth]{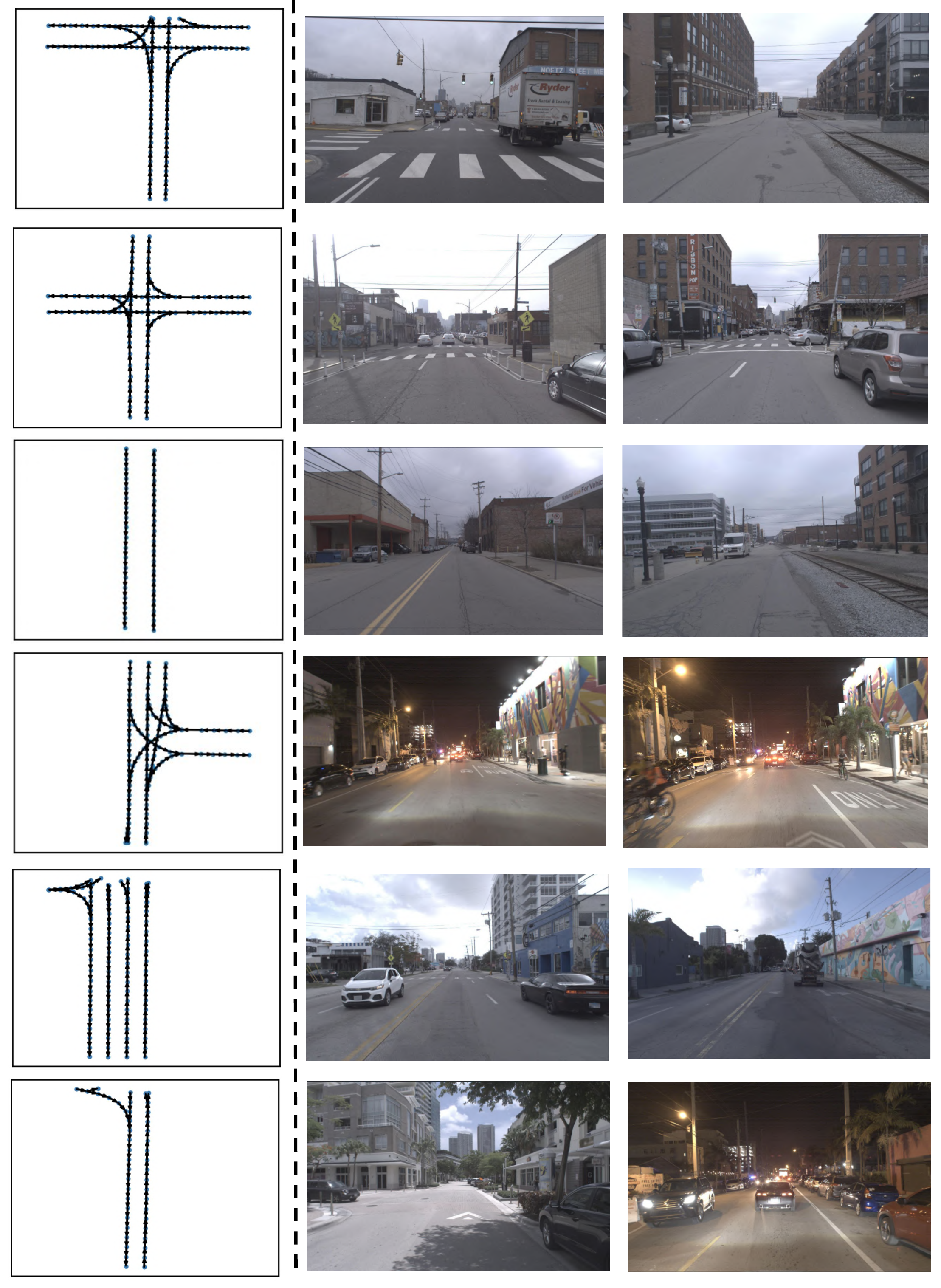}
  \caption{\textbf{Qualitative results for Map2Pix.} The goal is to retrieve ego-camera images given a street map. Such image retrieval may be useful for simulator-based training and validation of autonomous stacks. A single street geometry might retrieve multiple consistent, realistic imagery. 
  }
\label{fig:map2pix}
\end{figure}

We qualitatively evaluate the graph retrieval results in Fig. 6 in the main paper. %~\ref{fig:result}. 
Please see the caption for a detailed description, but generally speaking, we find \textit{Pix2Map} returns reasonable graphs similar to the ground truth. 
In Tab.~\ref{tab:update_exp_eval}, we evaluate the performance of map update and map expansion in two cities (Pittsburgh and Miami). We do so by comparing expanded/updated maps with ground-truth maps using metrics. As shown above, map expansion to novel areas is harder than updating previously-seen areas.

\subsection{Localization} 
\label{sec:localization}
Furthermore, our method demonstrates great visual localization ability based on visual and geometric understanding. We use the cosine similarities of retrieved graphs to generate a heatmap of possible ego-vehicle locations over a city-level map, showing the locations where their corresponding graphs are assigned a high likelihood, relative to the ground truth location shown as a red dot. While the ground truth is usually assigned a high likelihood, which indicates the promising performance of localization ability, the distribution becomes less sharp with respect to position when farther away from intersections. See Fig.~\ref{fig:localization_heatmap} for more details.

\subsection{Map2Pix} 
\label{sec:map2pix}
We further show that it is also possible to retrieve egocentric camera data using street maps. Such techniques could be used in the future to synthesize virtual worlds consistent with the query road geometry. We provide a few example image retrievals in Figure~\ref{fig:map2pix}, visualizing the front views of the top $K=2$ images for each street map. As can be seen, the retrieved images correspond to rough geometric layouts encoded in the query graphs.

% {\small
% \bibliographystyle{ieee_fullname}
% \bibliography{refs}
% }

% \end{document}

\end{document}